\documentclass[letterpaper, 10 pt, conference]{ieeeconf}  % Comment this line out if you need a4paper

\IEEEoverridecommandlockouts                              % This command is only needed if 
\usepackage{soul}
\usepackage[tracking=true]{microtype}
\usepackage{amsmath} % assumes amsmath package installed
\usepackage{amssymb}  % assumes amsmath package installed
\usepackage{graphicx}
\usepackage{epsfig} % for postscript graphics files
\usepackage{mathptmx} % assumes new font selection scheme installed
\usepackage{times} % assumes new font selection scheme installed
\usepackage{amsmath} % assumes amsmath package installed
\usepackage{amssymb}  % assumes amsmath package installed
\usepackage{xspace}
\usepackage{float}
\usepackage{afterpage}
\usepackage{url}

\usepackage[dvipsnames]{xcolor}
\usepackage[font={footnotesize}]{caption}
\usepackage{booktabs} % for professional tables
\usepackage{tikz}
\usetikzlibrary{positioning,shapes.geometric,arrows,fit,shapes.symbols,svg.path,tikzmark,calc}
\usepackage{textcomp}
\usepackage{listings}
\usepackage{subcaption}
\pgfdeclarelayer{background}
\pgfdeclarelayer{foreground}
\pgfsetlayers{background,main,foreground}
\usepackage{siunitx}
\usepackage{adjustbox}
\usepackage{array}
\usepackage{multirow}

\usepackage{siunitx}

\let\labelindent\relax %https://tex.stackexchange.com/questions/170772/command-labelindent-already-defined

\usepackage[inline]{enumitem} % for inline lists, e.g. \begin{enumerate*}[label=\roman*)]  See https://tex.stackexchange.com/questions/146306/how-to-make-horizontal-lists

\usepackage[backend=biber,
            url=false,
            isbn=false,
            doi=false,
            backref=false,
            style=ieee,
            natbib=true,%compatibility aliases
            mincitenames=1,
            maxcitenames=1,
            minbibnames=12,
            maxbibnames=12,
            citestyle=numeric-comp,
            sorting=none,%none
            block=none]{biblatex}
\renewcommand{\bibfont}{\small}
\newcommand{\algname}{Robo-DM\xspace}

\newcommand{\oxe}{OXE\xspace}

\newcommand{\numberedparagraph}[1]{\textit{#1:} }

\title{\LARGE \bf
Robo-DM: Data Management For Large Robot Datasets
}

\author{$^\dagger$Kaiyuan Chen$^{{1}}$,  Letian Fu$^{1}$,  David Huang$^{1*}$, Yanxiang Zhang$^{1*}$, Lawrence Yunliang  Chen$^{1}$, Huang Huang$^{1}$, \\ Kush Hari$^{1}$,  Ashwin Balakrishna$^{2}$, Ted Xiao$^{2}$, Pannag R Sanketi$^{2}$, John Kubiatowicz$^{1}$, Ken Goldberg$^{1, 3}$ 
\thanks{$^{1}$University of California, Berkeley}%
\thanks{$^{2}$Google Deepmind}%
\thanks{$^{3}$Department of Industrial Engineering and Operations Research}%
\thanks{$^\dagger$For correspondence and questions: {kych@berkeley.edu}}
}

\begin{document}

\maketitle
\thispagestyle{empty}
\pagestyle{empty}

\begin{abstract}
    % Roboticists conjecture the Vision-Language-Action (VLA) models using Transformer Networks could perform analogously to Large Vision Language models if trained on sufficient examples.  
Recent results suggest that very large datasets of teleoperated robot demonstrations can be used to train transformer-based models that have the potential to generalize to new scenes, robots, and tasks. 
 % achieving state-of-the-art
% performance on robotics multi-task settings
% \eric{talk about Open X, DROID, }
However, curating, distributing, and loading large datasets of robot trajectories, which typically consist of video, textual, and numerical modalities - including streams from multiple cameras - remains challenging.
% For instance, distributing the \oxe datasets via the cloud costs the data host hundreds of dollars per full download. 
% Existing solutions are over-engineered that requires extensive heuristics on how data is collected, stored and shared, and sometimes falls short in usability, reusability and efficiency. 
% To scale to the expected scale with potentially thousands of times of the current scale, a \textit{sustainable} growth is required. 
We propose \algname, an efficient open-source cloud-based data management toolkit for collecting, sharing, and learning with robot data. 
With \algname, robot datasets are stored in a self-contained format with Extensible Binary Meta Language (EBML). %Inspired by Matroska Video files (MKV), \algname preserves the original time information to automatically align heterogeneous data streams and transparently encodes vision data with video compression. 
\algname can significantly reduce the size of robot trajectory data, transfer costs, and data load time during training.
Compared to the RLDS format used in \oxe datasets, \algname's compression saves space by up to 70x (lossy) and  3.5x (lossless). 
\algname also accelerates data retrieval by load-balancing video decoding with memory-mapped decoding caches. 
Compared to LeRobot, a framework that also uses lossy video compression, \algname is up to 50x faster when decoding sequentially. 
% In fine-tuning Octo, a transformer-based robot policy with 73k episodes with RT-1 data, \algname incurs 2.6\% increase at training performance loss.
We physically evaluate a model trained by \algname with lossy compression, a pick-and-place task, and In-Context Robot Transformer. \algname uses 75x compression of the original dataset and does not suffer reduction in downstream task accuracy. 
% \eric{2.6 is the current number with incomplete runs. Need to be experimented with different hyperparameter}
Code and evaluation scripts can be found on website \url{https://github.com/BerkeleyAutomation/fog_x}.

\end{abstract}

\section{Introduction}
 % \eric{@Professor, there are a few names that can potentially be authors or need to be acknowledged:  Karl Pertsch, Prof. Jeffrey Ichnowski, Quan Vuong, Prof. Joseph E Gonzalez,  Prof. Ion Stoica?}
 
% \eric{maybe: talk about growth rate of the data and model }
% Recent work\eric{cite} achieves state-of-the-art performance on robotics multi-task settings if trained with large robotics datasets. 
% leads to more robust and generalizable robotics models, 
Recent work  ~\cite{octo_2023, open_x_embodiment_rt_x_2023, brohan2023rt1, brohan2023rt2,  kim24openvla, khazatsky2024droid, fu2024icrt} suggests Vision-Language-Action models ~\cite{brohan2023rt2,octo_2023,kim24openvla} 
can enhance robot capabilities and generalization in handling multiple settings in diverse environments. A key ingredient for large model training is large and well-curated datasets of teleoperated robot demonstration trajectories such as the Open-X Embodiment (OXE) dataset~\cite{open_x_embodiment_rt_x_2023}.
However, the current management of robot data is inefficient~\cite{open_x_embodiment_rt_x_2023}. Each robot demonstration consists of sequences of actions and observations, making the learning samples much larger and of diverse structrure compared to the images or text tokens in VLMs~\cite{2023GPT4VisionSC, 2023gemini, liu2024visual, alayrac2022flamingo,driess2023palme} and LLMs~\cite{touvron2023llama,achiam2023gpt4}. 
% Existing solutions sometimes require over-engineering and extensive heuristics on how data is collected, stored and shared, which fall short in usability, reusability and efficiency.
At tera or even penta-byte scale, which is sometimes characterized as Big Data~\cite{sagiroglu2013big}, existing robot data storage methods can be inefficient.
% , and there is a need for better data compression to reduce storage and manage data throughput during training. 
% \ken{\st{In this work, we reconsider the robot data collection, distribution, and training pipeline and} Existing data storage methods are somewhat ad-hoc. 
% In this paper, 
We propose \algname, an efficient data format with a toolkit for robot data collection, management, and training. 
% An average robot demonstration data size in DROID dataset is 18MB, compared to 

% \eric{check the size of other fields such as video, saying how large per sample is, vs how large a trajectory is in robotics}
\begin{figure}
    \centering
    \includegraphics[width=0.85\linewidth]{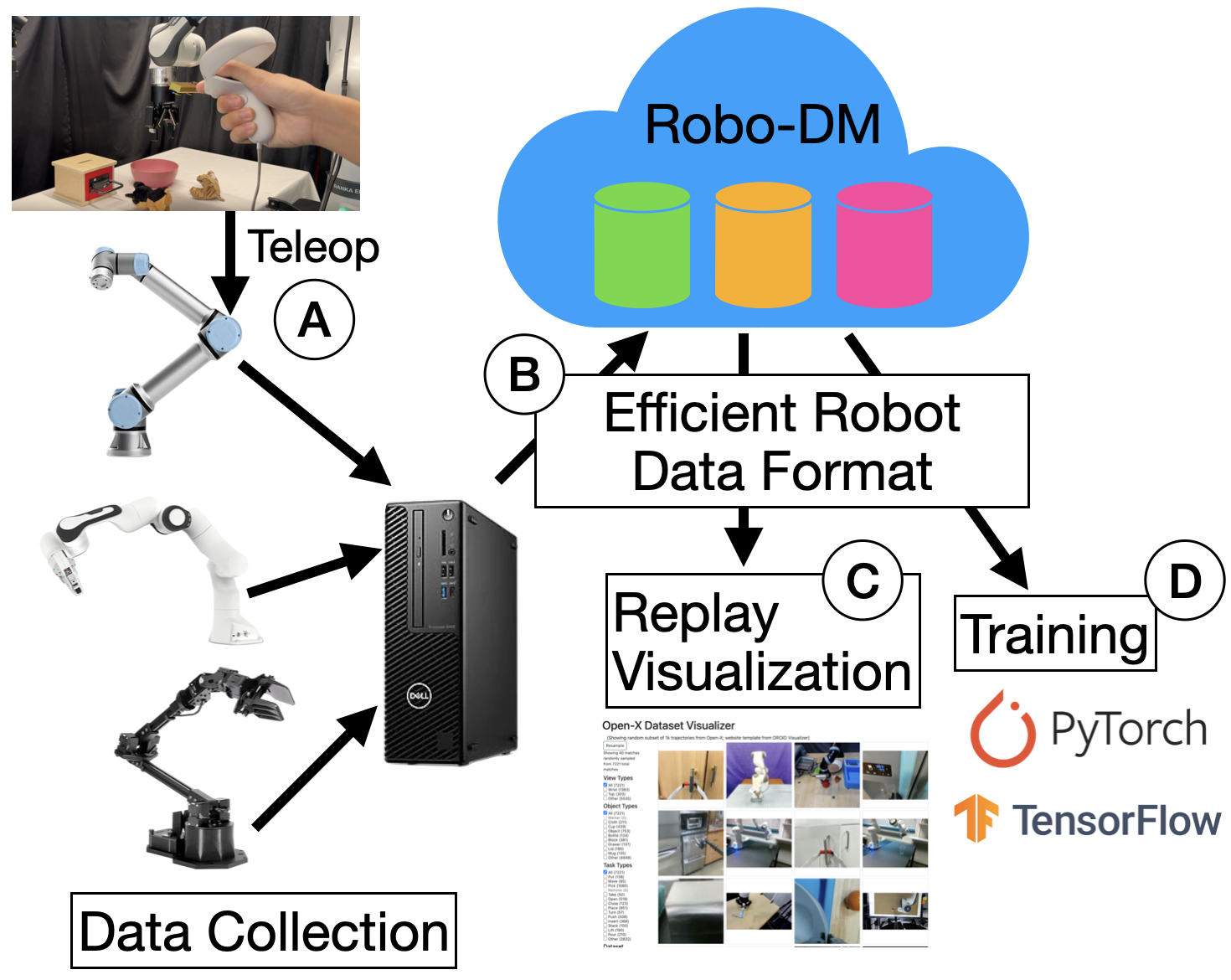}
    \caption{\algname can streamline robot data collection, management, and learning. (B) \algname uses a unified format for vision, language, and action that does not rely on assumptions about timestamps and data, and supports plug-and-play data collection to integrate with existing setups. (C) \algname can facilitate replay and visualization. (D) Existing training frameworks can load from \algname efficiently with minimal modification. }
    \label{fig:enter-label}
\end{figure}

A typical robot dataset includes a number of \textit{episodes}, a sequence of actions performed by an agent from a starting state to a terminal state. 
Each episode contains multiple sensor data streams in addition to language instructions and other metadata such as robot, task, environment, and control scheme specifications. 
% \ken{"features?" see fig. 2}
The size of a typical episode ranges from 1 MB to 400 MB, depending on the episode length, compression level, number of cameras, and camera resolution.  Data streams may be recorded at different sampling rates. Episodes are typically stored as a sequence of matrices; for example, data collection with DROID~\cite{khazatsky2024droid} automates data storage with  Hierarchical Data Format 5 (HDF5)~\cite{hdf5}, a format that supports hierarchical storage of matrices.  \oxe uses Reinforcement Learning Datasets (RLDS)~\cite{hussenot2021rlds}, an extension of Tensorflow Datasets (TFDS) to store reinforcement learning demonstrations. Storing image and sensor data directly in matrices is lossless but not space efficient. One emergent framework, LeRobot~\cite{cadene2024lerobot}, provides a platform to share robot models and datasets based on lossy video compression and HuggingFace datasets. However, its file structure is complex and loading is generally slower due to decoding.

We observe the following challenges in robot data collection and usage:
 
\numberedparagraph{(A) Transmission Efficiency} Distributing robotics datasets is costly. 
Cloud service providers, such as Google Cloud Platform (GCP) and Amazon Web Services (AWS), charge the \textit{data host} for both data storage and outbound data transfers. The cost of transferring data often exceeds the cost of storing it. 
For example, storing  8.9 TB of Open-X data on Google Cloud costs 172 US dollars per month, but \emph{every full download} costs between 172 US Dollars and 1,540 US dollars. \footnote{The rate is calculated with the egress network traffic pricing in Google Cloud Platform (GCP), where the Open-X-Embodiment dataset is hosted. We use the size of Open-X v1.1 dataset with 8,964 GB in total. The rate differs by the downloading source and destination region. The rate does not consider retransmission of lost packets, so the actual cost is higher than the estimation. } 
Directly training with cloud storage requires repeated downloads if the local storage cannot store the full dataset, further increasing network traffic and cost to the \textit{data host}. 
Thus, improving data compression and transmission efficiency can reduce host costs and encourage public sharing of datasets. 
% Even well funded companies benefit from reducing data hosting cost to maintain public accessible data and resources. 

% \numberedparagraph{(B) Reusability} Typically, robotics training data is collected for a specific task. In a VLA data collection setup, not all the sensors will be used for the task and sometimes only task-specific features are retained. This makes it challenging to reuse the data for different tasks with varying feature needs. \eric{Ideally, we want a fixed setup that mounts all the sensors available, so that we don't need to recollect the number if something changes.}

\begin{figure}
    \centering
    \includegraphics[width=\linewidth]{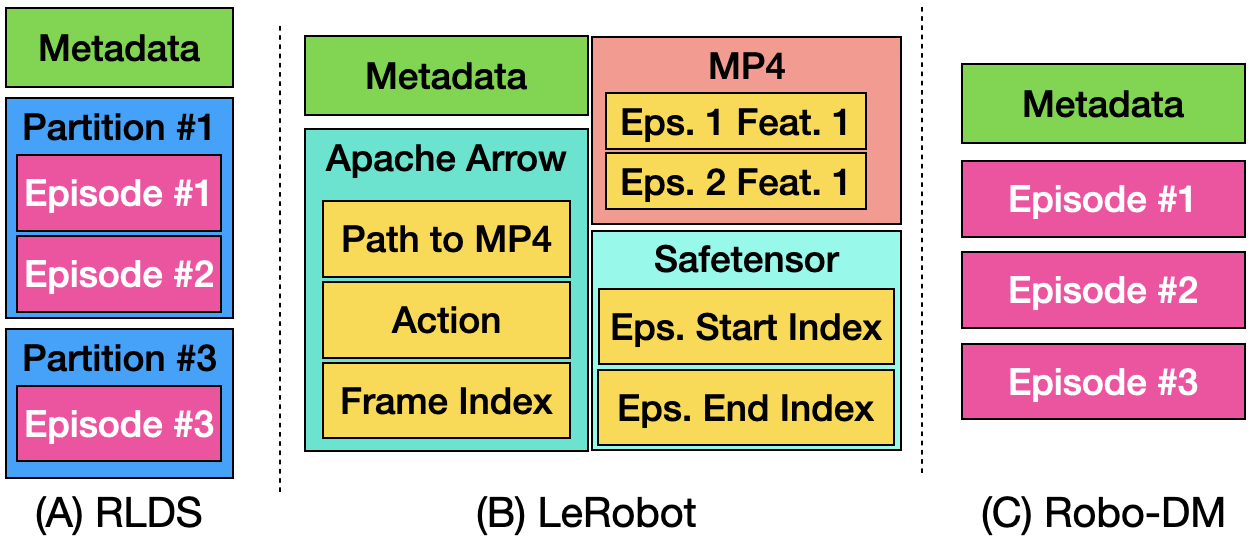}
    \caption{\textbf{A File Structure Comparison of RLDS, LeRobot and \algname} All formats include metadata, storing descriptive information such as authors and dataset summary. 
    (A) Reinforcement Learning Dataset (RLDS) stores episodes in partitions, where each partition is a Tensorflow Dataset Record file.  All streams in episode data are compressed matrices that can be directly loaded and trained in Tensorflow.    
    (B) LeRobot combines three formats for robot data. For vision data, it uses one MP4 per video stream in an episode, and uses HuggingFace Dataset (with Apache Arrow as backend\cite{arrow}) to store language and action streams and the path to the MP4 files.  
    It also uses  safetensors~\cite{safetensors} to store episode information. 
    All the streams are scattered: to extract an episode, the framework needs to query safetensors for episode information - which is used to find the rest of the non-video streams in the HuggingFace Dataset - and finally use the frame information from the HuggingFace Dataset to find the corresponding MP4 files for vision streams. (C) In \algname, robot data in all the episodes are stored and aligned in a self-contained format. To load an episode, one simply reads from \algname files and load as trainable matrices.}
    \label{fig:related:comparison}
\end{figure}

\numberedparagraph{(B) Usability and Simplicity}  Existing robot data frameworks impose restrictions on file structure, data layout, semantics, and alignment. 
% If the framework cannot meet a certain requirement, assumptions are layered over-engineering solutions, 
In particular, hybrid approaches rely on framework-specific assumptions to handle multiple formats simultaneously. Extending the current framework or migrating between frameworks can be challenging, resulting in complex structure and file organization. %Arching different formats require fully understand all the assumptions, and failure to follow lead to bugs and longer development cycle. 
Figure \ref{fig:related:comparison} shows a comparison of different storage formats.

\numberedparagraph{(C) Data Loading Performance} 
Large robot datasets are typically loaded into computationally training applications. 
In training, decoded frames are frequently reused and randomly accessed, and the decoded data is loaded on demand. 
Existing frameworks that use heavy compression sometimes lead to high computational resource utilization and interfere with the training performance. Thus, an efficient and perform ant data-loading framework should utilize available resources without contention.

We introduce \algname, an efficient cloud-based toolkit for collecting, sharing, and learning with robot data. 
% \pannag{Isn’t any framework going to impose similar assumptions along all of these dimensions? The goal is probably to argue that the assumptions you make are more general/less restrictive?}
\algname streamlines storage for vision, language, and action data via a unified container format with Extensible Binary Meta Language (EBML). 
% Inspired by Matroska Video (MKV) containers, \algname stores all time-aligned streams in a single container for multi-language and multi-device compatibility.
\algname efficiently orchestrates heterogeneous data streams, supporting flexible lossless compression and lossy compression for enhanced transmission efficiency. In prior work, LeRobot~\cite{cadene2024lerobot} empirically evaluates how lossy video compression parameters in FFmpeg affect robot policy accuracy. Octo is also pre-trained by compressing image frames in \oxe to lossy images~\cite{octo_2023}. 
\algname improves the data loading performance for training workloads, which requires repetitive data access by using memory-mapped caching for faster data retrieval and loading. 
Loading from cache and decoding are load-balanced to maximize the utilization of compute, memory and storage resources.
\algname requires minimal integration effort with existing frameworks. It supports plug-and-play data collection, training, replay, and visualization with mainstream frameworks, and can also be easily exported to other formats such as HDF5 and RLDS.

%This paper does not claim using lossy compression as a contribution. 

Experiments suggest that \algname can reduce the size of data by up to 70 times with lossy compression compared to how Open-X-Embodiment currently shares the dataset, and up to 50x faster than LeRobot, a framework that also uses lossy video compression to encode vision data. 
We fine-tune Octo, a transformer-based robot policy trained with an 800k Open-X-Embodiment dataset with 74k training episodes from RT-1. \algname reduces the dataset size by 4.39 times, while being 3.0 times faster in small batch size (data loading intensive) and does not introduce any slowdown in the training pipeline with large batch size (compute intensive).

This paper makes the following contributions: 
(1) an extension of EBML to define a container format that unifies time-based robot data storage;
(2) \algname, a framework with 6 new features using this container format;
(3) Experimental data that suggests \algname can significantly reduce dataset size, improve loading speed, and incur marginal training performance degradation.

\section{Related Work}

\textbf{Big Robot Data}
The robot learning community is actively building a number of open-source robot learning datasets~\cite{brohan2023rt1, kalashnikov2018qt, walke2023bridgedata}.
Recent work, such as Octo~\cite{octo_2023}, Open-VLA~\cite{kim24openvla}, are trained on large datasets such as RT-1~\cite{brohan2023rt1}, RT-2~\cite{brohan2023rt2},  Open-X-Embodiment~\cite{open_x_embodiment_rt_x_2023}, Distributed Robot Interaction Dataset (DROID)~\cite{khazatsky2024droid}.
Their initial results suggest training with large and diverse robotics datasets can enhance robot capabilities and generalization in handling multiple settings in diverse environments.  In this work, we present an efficient data pipeline for managing large and diverse robot datasets.

\textbf{Robot Data Frameworks} 
Existing frameworks for collecting, managing and storing robot data fall into the following three categories: 
(1) Serialized Log format that preserves timing information. This allows users to directly replay the data, e.g. with the official ROS2 tool, rosbag~\cite{quigley2009ros}.
(2) Matrix format that can be directly supported by training infrastructrue. For example, DROID~\cite{khazatsky2024droid} automates data storage with HDF5 Hierarchical Data Format (HDF5)~\cite{hdf5} and existing \oxe datasets use RLDS, an extension of Tensorflow Datasets (TFDS)~\cite{TFDS} that store and retrieve the interaction between an agent and an environment with observation, action and reward. Storing image and sensor data directly in matrices limits the capability of compression, and is thus not space efficient. 
(3) Hybrid formats that store different features in separate files and require assumptions on how different features are aligned and synchronized, such as LeRobot~\cite{cadene2024lerobot}, a platform to share robot models and datasets based on HuggingFace datasets.

\textbf{Cloud and Fog Robotics}
Fog Robotics~\cite{gudi2017fog} utilizes cloud and edge resources for robotics applications. Existing Fog and Cloud robotics focus on deployment of robotics applications, such as grasp planning~\cite{tanwani2019fog}, motion planning~\cite{ichnowski2020fog}, visual servoing~\cite{tian2019fog}, and human-robot interaction~\cite{gudi2018fog}. FogROS2~\cite{chen2021fogros} automates cloud compute resources for robotics, addressing issues such as connectivity~\cite{chen2023sgc}, latency~\cite{chen2024fogrosls}, and cost~\cite{chen2024fogrosconfig}.
We recognize the cost of the cloud required to distribute large robot datasets, and study 
how formats affect robotics learning in data collection, loading and management.

% \textbf{Video Compression and Containers} \eric{I think it could be fun to survey video containers - people typically don't know much about their differences}

\section{\algname Features}

Six novel features differentiate 
\algname from existing robot data frameworks:

\numberedparagraph{(1) Self-Contained Robot Data Storage} \algname uses a self-contained file format that integrates and stores heterogeneous robot data streams, ensuring all necessary data is consolidated within a single file.

\numberedparagraph{(2) Vision, Language, Action Data Orchestration} The format of \algname allows diverse binary robot data streams, including sensor data, environment specifications, language instructions, and kinematic controls.

\numberedparagraph{(3) Data Flexibility} 
\algname is extensible for new different data streams, compression algorithms and video encoding formats. For example, \algname enables users to flexibly choose from storing vision data as a sequence of serialized matrices, images, or encoding with lossy or lossless video codecs.
With \algname, one can record all the data  with original timestamps without resorting to heuristics on data alignment.

\numberedparagraph{(4) Efficient Dataset Size} 
% \algname significantly reduces the data required for transmission through compression. In Open-X-Embodiment, data is shared with trainable matrices, which leads to long loading time from the cloud and high network transmission cost. 
\algname efficiently encodes heterogeneous time-aligned streams.
It uses video compression to significantly reduce the size of file transfer. 

\numberedparagraph{(5) Data Loading Efficiency}
\algname efficiently loads data by caching decoded frames and balancing resource utilization across available hardware.

% \paragraph{Multi-Cloud Sharing and Loading} \algname supports directly loading from major cloud storage such as Google Cloud Platform and Amazon S3. 
% Because kinematic controls and environment specifications may run at different frequency as cameras, typically people need heuristics on how data is synchronized and used, such that all the features are of the same sampling frequency.
\numberedparagraph{(6) Simple Data Collection, Training and Visualization} 
\algname adopts a concise interface for data collection that is compatible with existing systems with minimal modification.  It integrates seamlessly with TensorFlow and PyTorch interfaces, enabling easy adoption.
It also allows for exporting of the collected data to existing state-of-the-art data storage frameworks, such as RLDS and HDF5. 
\algname supports replaying messages through Robot Operating System (ROS) 2, the de-facto standard for developing robotics applications. One can use off-the-shelf ROS2 tools such as rviz~\cite{kam2015rviz} or Foxglove~\cite{foxglove} to visualize the replayed streams.

\section{\algname Design}

\algname uses a unified and self-contained robot data format (Sec. IV.A) considering feature (1) (2) (3) in Sec. III. We design efficient storage and loading mechanisms (Sec. IV.B) for feature (4) and (5). Finally, \algname is integrated with existing frameworks for feature (6).

\subsection{Unified and Self-Contained Robot Data Format}
\label{sec:design:format}

\begin{figure*}
    \centering
    \includegraphics[width=0.75\linewidth]{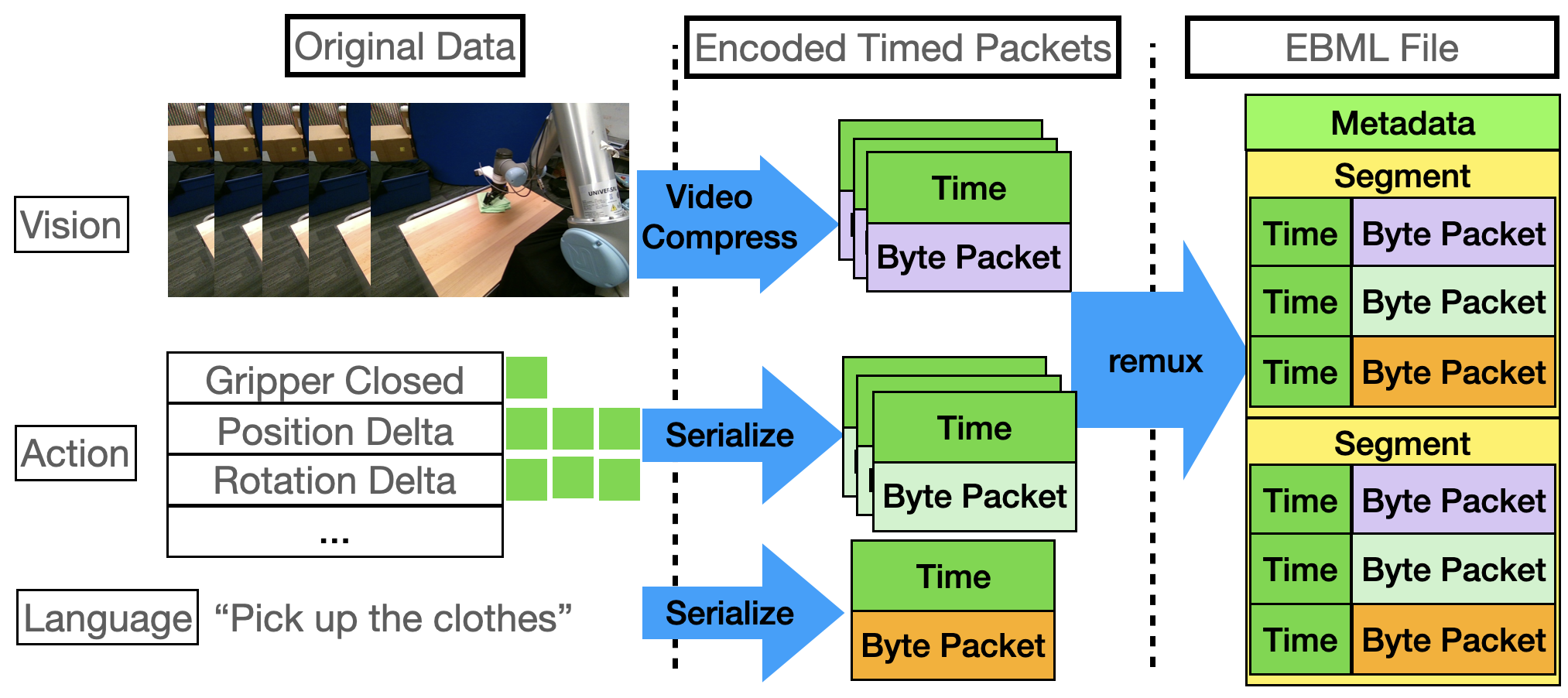}
    \caption{\textbf{How \algname stores an episode of robot data with vision, language and action data} \algname encodes vision, language and action data. For vision data, \algname uses video or image compression; language and action data are serialized into bytes. All the bytes are encapsulated with an intake timestamp. Then \algname multiplexes different streams of data into  a self-describing EBML file format. }
    \label{fig:design:mkv}
\end{figure*}

\algname uses Extensible Binary Meta Language (EBML) ~\cite{rfc8794} for data structuring. EBML is a versatile and extensible markup language that combines the flexibility of Extensible Meta Language (XML) with the efficiency of binary encoding. It organizes binary data elements in a hierarchical structure similar to XML, allowing for nested elements and coherent data management. This enables EBML to handle data streams from different sources within a single container, using self-describing elements that ensure compatibility and future extensibility. A notable application of EBML is in the MKV~\cite{mkv} video container format, which uses it to store multiple video and audio tracks, along with subtitles, in a time-aligned manner within a single container. 

Figure \ref{fig:design:mkv} illustrates how \algname encapsulates heterogeneous robot data streams. 
\algname compresses vision streams and serializes robot data into byte packets. A byte packet encapsulates the raw bytes and descriptive information, such as timestamp and stream information.
To efficiently replay the data and keep the relative timing information between data streams with different frequencies, all the data packets are stored with a relative timestamp to the beginning of the episode.
\algname extends MKV to store robot data to ensure the synchronization of multiple streams on vision, language, and action within the same container.

% MKV uses this nature of EBML to store multiple streams of the same video, audio and subtitles, so that the video can play at different resolutions, languages and devices. 
% \algname uses this property to store data streams of vision language and action data.

% \eric{timestamp is labeled}

\textbf{Data Collection and Post-Processing}
Compression can be computationally intensive. To prevent interference with the data collection process, \algname uses its file format flexibility to first store all data in raw serialized form. 
After the data collection is finished,  \algname iterates through the collected data, transcodes data that requires compression and re-arranges the collected data (remux) to arrange the data packets. 
Because training applications sometimes access the episode at a given time frame, \algname groups time-aligned data streams together. On querying a specific frame, metadata is used to identify the related segments and decode the video starting from the latest keyframe before the start of the slice. All the decoded trajectories are cached to speed up future accesses.

% This capability makes EBML especially suitable for applications where multiple, need to be stored, accessed, and processed efficiently, as it supports the seamless integration and management of various data types within a unified container.

% \eric{do we need more information about how different streams are aligned?}
% \eric{maybe in the figure, show how ebml align different streams and features}

\subsection{Transmission-Efficient Storage, Retrieval and Loading}
\label{sec:design:loading}
\textbf{Transmission-Efficient Compression} 
\algname unifies heterogeneous data streams that require different mechanisms for compression and serialization. 
Because \algname naturally supports byte streams, it is agnostic to mainstream byte compression algorithms and video encoders.
For vision data, three channels (red, green, blue) can be compressed with off-the-shelf video compression algorithms, such as H.264~\cite{h264}, H.265~\cite{h265}, AV1~\cite{av1}. For large matrices that require full precision, such as stereo depth images, users can alternatively choose to compress them with lossless compression algorithms such as FFV1~\cite{loc_ffv1, rfc9043}.

\textbf{Efficient Decoding Cache} For sequential access patterns, compression-based algorithms can reduce space usage by decoding all frames in order. When training, decoded frames are frequently reused and randomly accessed, and the decoded data is loaded on-demand. \algname amortizes the random access patterns using memory-mapped files (mmap)~\cite{mmap, hdf5}. Mmap creates a new mapping in the virtual address space of a process to a cache file.
If a slice of data is used, only the portions of the file that are actually used are brought into memory, conserving both I/O bandwidth and physical memory. 

\textbf{Load Balancing For Decoding and Decoding Cache} \algname automates the choice of computationally heavy decoding, loading directly cache in memory, and loading the decoded matrices from disk. To prevent overusing a single resource, \algname estimates the potential latency of accessing the data and dynamically balancing the access. Specifically, if the memory resources are underutilized and a prior decoded matrix is available, this means the decoded data is likely in physical memory without being cached to the disk by mmap, and \algname can directly use the decoded cache. In contrast, if the memory is full, cache miss is frequent and the data is not frequently accessed, \algname does not load from cache, and directly decodes the video data instead.

\begin{table*}[h]
\centering
\resizebox{\linewidth}{!}{
\begin{tabular}{l|ccc|rrrrr}
\toprule
      &\multicolumn{3}{c|}{\textbf{Dataset Description}} & \multicolumn{5}{c}{\textbf{Total Dataset Size (GB)}} \\
\textbf{Dataset} &  &  & \textbf{Avg. Frames}  & \textbf{Original} & & \textbf{\algname-} &  & \\
  & \textbf{\# Image Streams}     &   \textbf{Resolution}   &   \textbf{per Episode}   & \textbf{RLDS} & \textbf{HDF5}  & \textbf{Lossless} &  \textbf{LeRobot} & \textbf{\algname}  \\
\midrule
\textbf{Bridge}        &  1 RGB &   (480, 640) &  34  &    387.49 (73x)  &  779.24 (147x) &  114.63 (22x) & 16.34 (3x) & \textbf{5.31 (1x)}\\

\textbf{Cable Routing} &  3 RGB          &  (128, 128) & 25  &    4.67 (18x) & 7.38  (28x)   & 1.67 (6x) & 0.36 (1.4x) & \textbf{0.26 (1x)} \\
\textbf{Door Opening}  &  1 RGB          &   (720, 960) &  42    &  7.12 (71x)  & 35.35  (354x)  & 2.89 (29x) & 0.38 (4x) & \textbf{0.10 (1x)}\\
\textbf{AutoLab UR5}   &  2 RGB, 1 Depth &   (480, 640)      & 97   &  76.39 (23x)   & 258.33 (88x) & 23.45 (7x) & (-) & \textbf{3.26 (1x)}\\ %2.22 
% \textbf{RT-1}          &  1 RGB          & (320, 480) & 73,499   & 111.06 &  &  &  - & -  \\
\bottomrule
\end{tabular}
}
\caption{\textbf{Dataset information and Size Comparison with Different Formats in Gigabytes (GB)}. Compression ratios differ by the number of image streams and resolution. \algname and LeRobot use lossy compression, while the rest are lossless. Both LeRobot and \algname use AV1 codec with 30 Constant Rate Factor (CRF), a factor that balances compression and decoded video quality. These parameters are suggested by LeRobot video benchmark~\cite{lerobot_benchmark}. (-) LeRobot omits depth stream and some action streams at its conversion from RLDS~\cite{hussenot2021rlds}.}
\label{tab:dataset:spec}
\end{table*}

% \begin{table}[h]
% \centering
% \resizebox{\linewidth}{!}{%
% \begin{tabular}{l|rrr}
% \toprule
%   & \textbf{\begin{tabular}[c]{@{}c@{}} Cable Routing \end{tabular}} & \textbf{Bridge} & \textbf{\begin{tabular}[c]{@{}l@{}} Door Opening\end{tabular}} \\
% \midrule 
% HDF5  & 4.86 (4.4x) & 29.91 (6.8x) & 79.54 (13.8x) \\
% RLDS   & 3.23 (2.9x) & 15.58 (3.5x) & 16.76 (2.9x)  \\
% \algname & \textbf{1.10 (1.0x)} & \textbf{4.37 (1.0x)} & \textbf{5.78 (1.0x)} \\
% \bottomrule
% \end{tabular}%
% }
% \caption{\textbf{Average Episode Size by Dataset and Lossless Format in Megabytes (MB) Compared to \algname} \algname uses lossless FFV1 codec to encode images. } 
% \end{table}

% \begin{table}[h]
% \centering
% \resizebox{\linewidth}{!}{%
% \begin{tabular}{l|cccc}
% \toprule
%   & \textbf{\begin{tabular}[c]{@{}c@{}} Cable Routing \end{tabular}} & \textbf{Bridge} & \textbf{\begin{tabular}[c]{@{}l@{}} Door Opening\end{tabular}} \\
% \midrule
% LeRobot   & 0.68 (3.8x) & 0.31 (1.6x)   & 0.88 (4.0x)   \\
% \algname  & \textbf{0.18 (1.0x)} & \textbf{0.20 (1.0x)}   & \textbf{0.22 (1.0x)}   \\
% \bottomrule
% \end{tabular}%
% }
% \caption{\textbf{Average Episode Size by Dataset and Format in Megabytes (MB) of lossy compression}. LeRobot omits depth stream and some action streams at its conversion from RLDS. Both LeRobot and \algname use AV1 codec with 30 Constant Rate Factor (CRF), a factor for compression and decoded video quality.} 
% \end{table}

\subsection{Integration with existing Frameworks}
\label{sec:design:collection}

% \begin{listing}[t]
% \inputminted[xleftmargin=20pt,linenos,fontsize=\scriptsize,escapeinside=||]{python}{listing/launch_example.py}
% \caption{\textbf{Code Example} \algname adopts a minimalist data collection, loading and exporting interface that can be easily integrated with existing frameworks.}
% \label{lst:code_example}
% \end{listing}

\textbf{Data Collection Interface}  
In order to integrate with custom data collection software stacks, \algname uses a concise programming interface for data collection. \algname data collection library infers time and the data type from the input vision, action and language data. Due to the simplicity in \algname's data storage format, the data collection library introduces minimal code complexity to the overall custom data collection software stack.

\textbf{Plug-and-Play Data Collection and Visualization}
\algname supports integration with ROS2-enabled setups to collect data in a plug and play manner. In ROS2, computational modules, \emph{nodes}, can be deployed on different machines. ROS2 provides an off-the-shelf tool, \emph{rosbag}, to capture data streams from sensors, logs, and various topics during robot operation. 
\algname supports transcoding from and exporting robot data to rosbag, with all the timing information recorded. Rosbags also can be directly replayed in ROS2. The ROS2 community provides a number of frameworks, such as  rviz~\cite{kam2015rviz}, and Foxglove~\cite{foxglove} from the open source community, a browser-based tool that enables visualization of ROS 2 topics.
Besides replaying videos, these visualizers also support visualization in 3D, which is helpful for action data such as robot state and motions.

\textbf{Data Loading Interface}
To support existing training frameworks with minimal modification, \algname supports accessing robot data in the same way as accessing typical HDF5 files (shown in Listing ~\ref{lst:code_example}).
\algname supports converting robot data to other state-of-the-art formats, such as HDF5 and Tensorflow dataset.

\section{Evaluation}

Our experiments consider three questions: 
(1) How does \algname's training data loader compare with state-of-the-art data loaders?
(2) How does \algname work with training workloads in terms of data loading speed, space saving, and training performance? 
(3) Does \algname preserve the policy performance? 

\textbf{Setup} We evaluate \algname with a standard workstation setup: Intel i9-13900K Processor with 96GB RAM and NVidia 4070 Ti Super GPU. The workstation is equipped with 6TB NVMe M.2 SSD with the reading throughput up to 5000 MB/s and writing throughput up to 2500 MB/s. It connects Internet with a 1 Gbps Ethernet connection that can download from Open-X-Embodiment Google Cloud Bucket with 10 Mbps.
We make sure the batch can fit in RAM without swap space. 
The video streams in \algname are decoded with CPU without specialized GPU or additional hardware decoder.

\subsection{Data Loading Benchmarks with Open-X-Embodiment}

% \begin{table}[h]
% \centering
% \resizebox{\linewidth}{!}{%
% \begin{tabular}{l|cccc}
% \toprule
%  & \textbf{\begin{tabular}[c]{@{}c@{}}AutoLab UR5\end{tabular}} & \textbf{\begin{tabular}[c]{@{}c@{}} Cable Routing \end{tabular}} & \textbf{Bridge} & \textbf{\begin{tabular}[c]{@{}l@{}} Door Opening\end{tabular}} \\
% \midrule
% HDF5    & 281.55 (11.01x) & 4.86 (4.42x) & 29.91 (6.80x) & 79.54 (13.76x) \\
% RLDS    & 87.30 (3.41x)  & 3.23 (2.94x) & 15.58 (3.54x) & 16.76 (2.90x)  \\
% \algname-Lossless & \textbf{25.57 (1x)} & \textbf{1.10 (1x)} &\textbf{ 4.40} (1x) & \textbf{5.78 (1x)} \\
% \midrule
% LeRobot & 2.41 (0.09x)*   & 0.68 (0.62x) & 0.31 (0.07x)*   & 0.88 (0.15x)   \\
% \algname     & \textbf{1.85 (0.07x)} & \textbf{0.23 (0.21x)} & \textbf{0.22 (0.05x)}   & \textbf{0.36 (0.06x)}   \\
% \bottomrule
% \end{tabular}%
% }
% \caption{\textbf{Average Episode Size by Dataset and Format in Megabytes (MB) and Number of Times Space Compared to \algname-L}. The numbers in parentheses represent the relative file size compared to \algname-L as the baseline (1x). (*) LeRobot omits depth stream at its conversion from RLDS.}
% \label{tab:trajectory_size}
% \end{table}

\begin{figure}
    \centering
    \includegraphics[width=\linewidth]{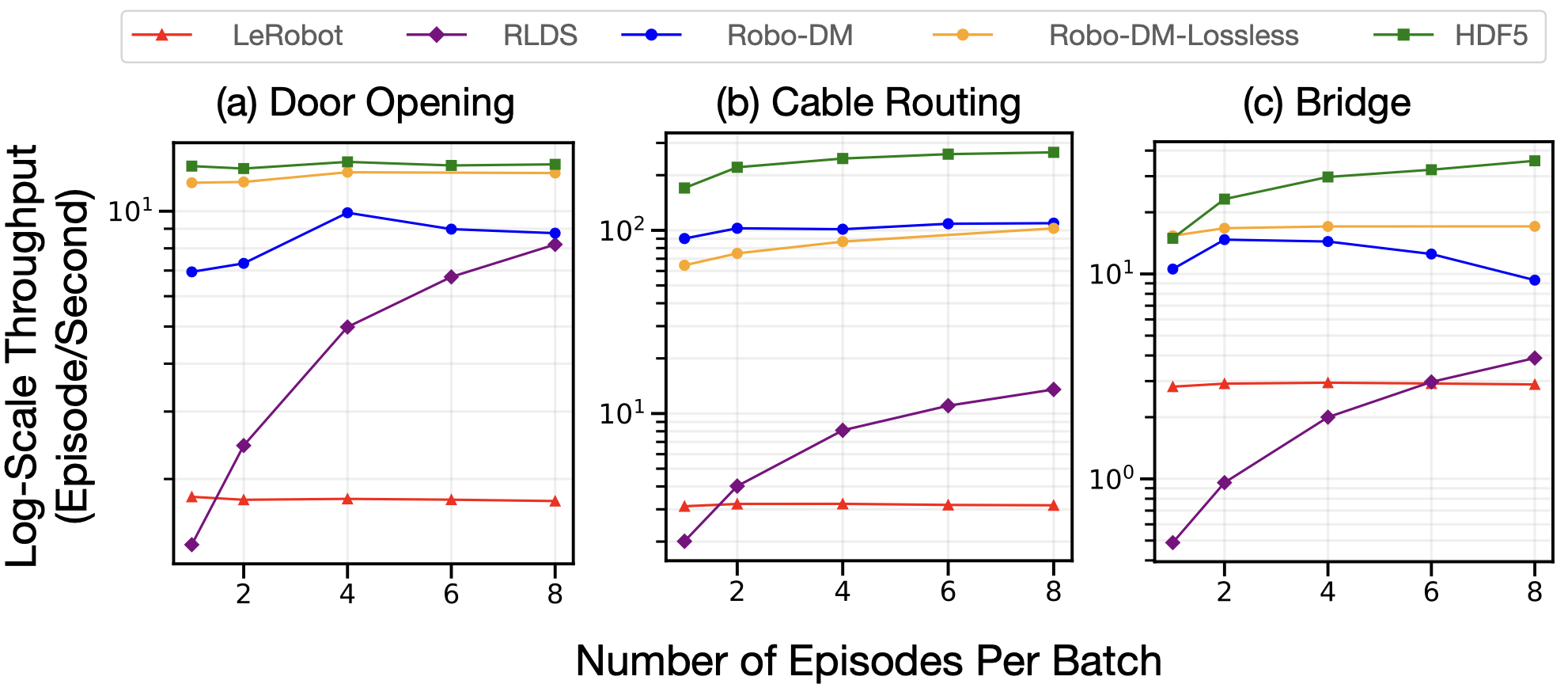}
    \caption{\textbf{Episode Per Second Throughput of \textcolor{blue}{\algname} on Three \oxe datasets with Different Characteristics} We compare \algname with baseline data loading Methods \textcolor{purple}{RLDS}, \textcolor{ForestGreen}{HDF5} and \textcolor{red}{LeRobot}. Complete episodes are loaded concurrently as a batch, and we record the average throughput with 200 batches.  }
    \label{fig:eval:throghput}
\end{figure}
\begin{figure}
    \centering
    \includegraphics[width=\linewidth]{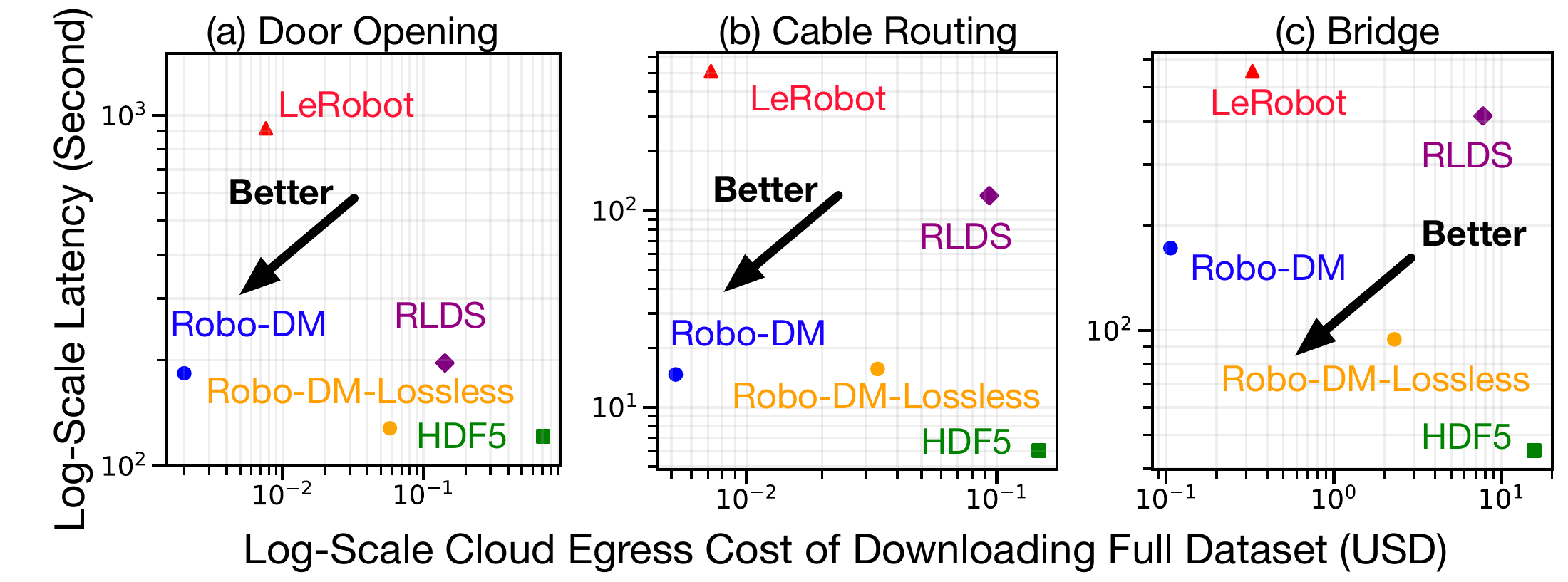}
    \caption{\textbf{Concurrent Loading Latency with respect to Episode Size of \textcolor{blue}{\algname}} We compare \algname with baseline data loading Methods \textcolor{purple}{RLDS}, \textcolor{ForestGreen}{HDF5} and \textcolor{red}{LeRobot}. Complete episodes are loaded concurrently as a batch, and we record the average latency of  200 batches with batch size 8 episodes. We use the lowest GCP cost of 0.02 US Dollars (USD) per GB.}
    \label{fig:eval:cost}
\end{figure}

% setup 

We evaluate the data loading performance of \algname with a number of exemplar datasets from Open-X-Embodiment (OXE).  In the experiments, we concurrently load multiple entire episodes into memory, and we explicitly cast the data into in-memory numpy arrays. We measure the latency of issuing a number of concurrent reads (i.e. a batch) to the time that all the episodes are loaded. For each run, we measure the average latency over 200 data loads. 

\textbf{Datasets} We use 
\begin{enumerate*}
    \item \emph{Bridge}~\cite{walke2023bridgedata}: two WidowX arms interact with household environments including kitchens, sinks, and tabletops. Skills include object rearrangement, sweeping, stacking, folding, and opening/closing doors and drawers. In the dataset,there are 4 RGB streams and 1 depth stream with 25,460 training episodes. 
    \item  \emph{UC Berkeley Cable Routing}:~\cite{luo2024multi} one Franka robot arm routes a cable through a number of tight-fitting clips mounted on the table with  1,482 training episodes. 
    \item \emph{NYU Door Opening}:~\cite{pari2021surprising} A Hello Stretch robot opens cabinet doors for a variety of cabinets  with 435 training episodes. 
    \item \emph{Berkeley AUTOLab UR5 ~\cite{BerkeleyUR5Website}}: A UR5 robot arm pick-and-place of a stuffed animal between containers, sweeping a cloth, stacking cups with 896 training episodes. 
\end{enumerate*}

\textbf{Baselines} We compare \algname with the following baselines 
\begin{enumerate*}
    \item \emph{RLDS}~\cite{hussenot2021rlds} Open-X-Embodiment is stored and shared in RLDS format. In the evaluation, we directly download and load the datasets with official instructions.
    \item \emph{LeRobot}~\cite{cadene2024lerobot} We convert Open-X-Embodiment datasets in LeRobot datasets with the provided official script. Some features in Open-X-Embodiment are omitted in the conversion. We sequentially extract episodes suggested by the example instructions. 
    \item \emph{HDF5}~\cite{hdf5} We use \algname to convert Open-X-Embodiment datasets to HDF5 formats. Since one HDF5 file per trajectory, we implement pre-fetch buffer and pytorch loader with the same setup as \algname.  
\end{enumerate*}
We use a pre-fetch buffer of 50 episodes.

\textbf{Episode Size} 
Table \ref{tab:dataset:spec} shows that 
\algname significantly reduces file size (18x, 73x, 23x and 73x) per episode compared to the RLDS, a format in which these datasets are originally stored and shared.
The episode size reduction leads to high accessibility to large robot datasets, transmission efficiency, and cost efficiency, shown in Figure \ref{fig:eval:cost}.

\textbf{Loading Latency}
Figure \ref{fig:eval:throghput} compares the throughput difference of \algname compared against LeRobot, RLDS, and HDF5. The lossless version of \algname has similar throughput as \algname. It is faster than LeRobot by 33x, 20x and 5x. \algname is slower than HDF5 because the HDF5 data is uncompressed and loaded in high disk throughput.

\textbf{Limitation}
Because \algname extensively uses RAM as a decoding cache to prevent repetitive decoding of the data, it leads to higher RAM usage and potentially degrades the performance when the per-episode data is large. For example, for the bridge dataset, we see \algname reduces the overall throughput when the batch size increases. This may lead to performance degradation for decoding for finer granularity, such as only sampling one frame for each episode.

% Specifically, AutoLab UR5 dataset uses multiple higher resolution with (640x480) RGB streams and 1 depth stream, \algname needs to frequency resort to video decoding and thus leads to similar performance pattern as LeRobot. 

\subsection{Case Study: Fine-tuning Octo with \algname}

% \begin{table}
% \centering
% \begin{tabular}{c|c c }
% \toprule
% \textbf{Batch Size} & \textbf{RLDS} &  \textbf{\algname }\\
% \midrule
% 1 &  0.024 & \textbf{0.008} \\
% 2 &  0.028 & \textbf{0.026} \\
% 16 &  0.071 & 0.071 \\
% % 32 & 0.097 & 0.097 \\
% 64 & 0.185 & 0.185 \\
% \bottomrule
% \end{tabular}
% \caption{\textbf{Per-Iteration Latency of Fine-Tuning Octo policy with different batch sizes} Small batch size is data loading intensive, and \algname has higher data loading throughput. When it comes to large batch size, the compute power is the bottleneck, and \algname does not lead to performance degradation. \eric{data loading time - model loading time - warm up } } 
% \label{tab:latency_comparison}
% \end{table}

Octo~\cite{octo_2023} is a transformer-based robot policy trained on 800k robot episodes from Open-X-Embodiment.
We fine-tune the pre-trained Octo-small model with 25.6M trainable parameters. We fine-tune the entire model conditioned with both images and language instructions. For each configuration, we train with 50,000 iterations and measure the per-iteration average latency.

\textbf{Dataset Compression} We use RT-1~\cite{brohan2023rt1} dataset, a dataset containing 73,499 episodes. The dataset involves picking, placing, and moving 17 objects with Google Robot. The dataset contains 1 RGB video stream with resolution (320, 480). 
The original dataset is 111.06 GB. The final dataset size of \algname is 36.50 GB with 4.39 times size reduction. The reason why the size reduction is smaller than other datasets from Open-X-Embodiment is that the per-trajectory size is small, with 1.51 MB on average per trajectory in RLDS. \algname needs more space to store metadata for seeking and decoding. 

\textbf{Training Performance} We run the training workload with batch size 64.  Dataloader in Octo loads from Tensorflow dataloader and \algname and lead to similar data loading latency (0.02 seconds) per iteration and overall training latency (0.10 seconds) per iteration. In validating the effect of lossy compression to the training outcome, we use lossless dataset for validation. The final image-conditioned Mean Squared Error of validation dataset is 1.86 with original lossless data and 1.91 with lossy data. Thus \algname leads to 2.6\% increase in validation loss with lossy compression. 

\subsection{Case Study: \algname with In-Context Robot Transformer Training}

\begin{figure}
    \centering
    \includegraphics[width=0.75\linewidth]{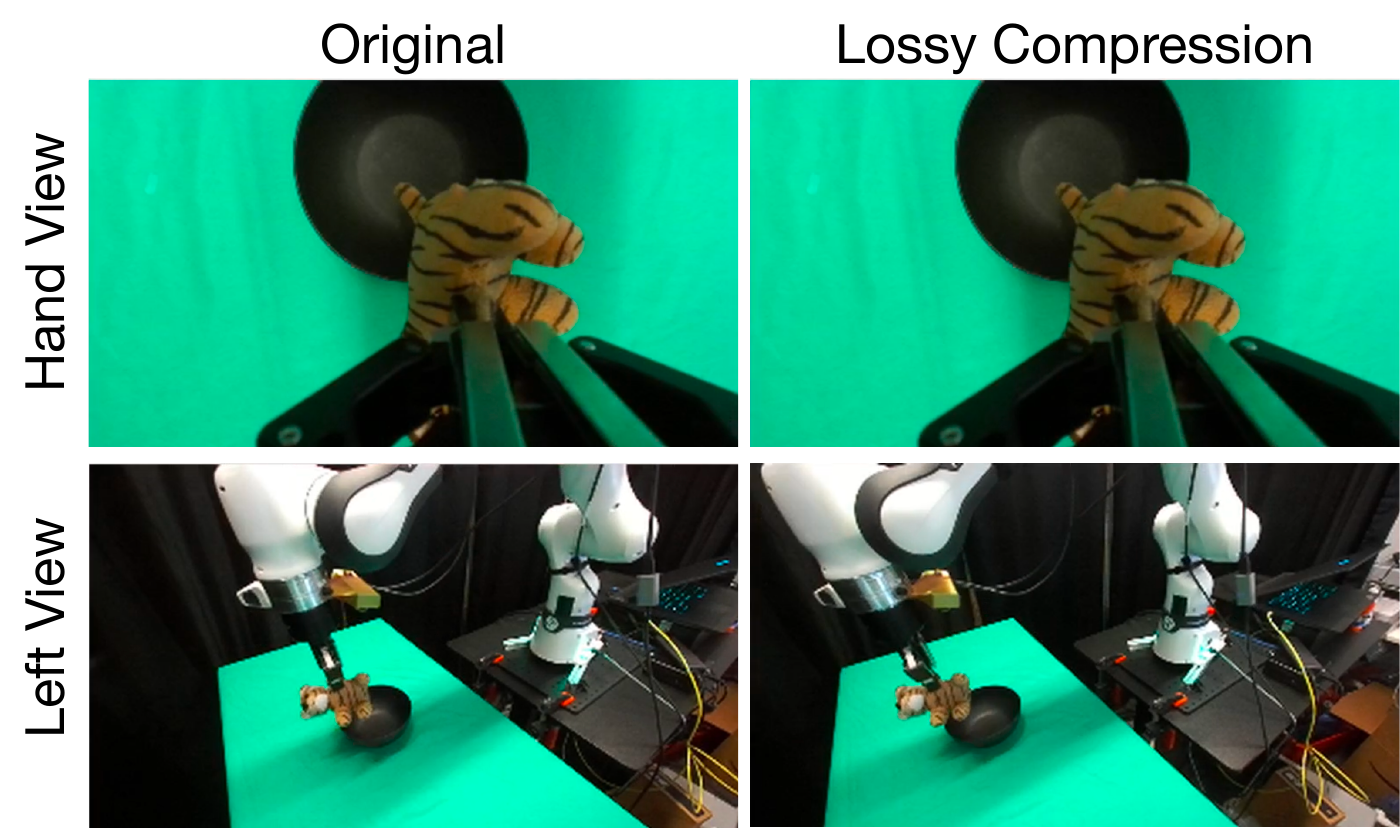}
    \caption{\textbf{ICRT Physical Experiment Setup with \algname} We setup ICRT to pick up a stuffed toy tiger and place it into a black bowl with a Franka Emika robot arm. The Figure shows the view from the left camera and wrist camera used for training, for both the original dataset and reconstructed images from \algname. }
    \label{fig:eval:tiger}
\end{figure}

\textbf{Task} We evaluate the training performance of \algname, hypothesizing the lossy compression of \algname, despite a high compression rate, could reduce the accuracy of the trained model. Thus, we evaluate a model trained with 335 human-demonstrated trajectories with the lossy compression of \algname. The trained model is tasked to pick up a stuffed toy tiger. Figure \ref{fig:eval:tiger} shows the task setup with the Franka Emika robot.

\textbf{Data} 
% number of traj, compression level to the original size, MSE to the original image, resolution \eric{need to confirm}. 
We collect 335 human-demonstrated trajectories with one hand camera and one left-side-view camera. All video streams are recorded at resolution (320, 180).
The trajectories were originally collected in HDF5 with gzip compression, with a total size of 5.8G. Stored in \algname's format, the dataset with lossless codec leads to 1.7G (3.41x space reduction), and the size of lossy compression is 77MB (75.3x space reduction).

\textbf{Model} We use the ICRT~\cite{fu2024icrt}, a transformer model that performs autoregressive prediction on sensorimotor trajectories. We train for 200 epochs with image brightness and contrast augmentation and a small proprioception noise ($\mathbf{N}(0, 0.01)$). 

\textbf{Results} We randomize the position of the stuffed toy tiger at different places on the tabletop. We evaluate with consecutive 15 trials on the model trained with lossy data.  The model is able to reliably identify the object, pick it up, and place it in a bowl with a 15 out of 15 success rate (100\%). 

\section{Conclusion}
In this paper we propose \algname, which includes a new format for robot data, and a toolkit for data collection, management, and loading. \algname significantly outperforms Open-X-Embodiment in terms of space saving. It also shows performant loading speed compared to LeRobot, a framework that also uses video compression.
In the task of fine-tuning Octo and policy training, \algname reduces dataset size with minimal training performance and accuracy degradation.

% In this work, we are mainly concerned with a usable and efficient file format, reaching comparable performance in training as other state-of-the-art methods. 
The file size reduction is mainly due to video compression. In future work, we will accelerate video compression and analyze the tradeoffs between parameters.
In the evaluation, we used the off-the-shelf video processing library, pyav \cite{pyav}, without GPU acceleration. Recent works such as Decord~\cite{decord} and GPU acceleration by Nvidia NVDEC~\cite{nvdec} are demonstrated to be faster than pyav. Also in future work, we will integrate and evaluate \algname with larger-scale of existing and prospective Open-X-Embodiment datasets.

\section{Acknowledgement}
This project benefited from discussions with Peter Schafhalter, Silvery Fu, and You-Liang Tan. This work is supported in part by donations from Google.

% \algname recognizes three tradeoffs that are task and hardware setup specific: 

% \textbf{Decoding Speed vs Data Loading Speed} \algname significantly reduces the amount of data transfer by XX times. In the evaluation, we explored two scenarios, NVMe SSD that has highest bandwidth that favors large data transfer, and cloud transfer with limited bandwidth that favors \algname. In the setting of hihg bandwidth transfer, \algname uses memory-mappable cache to reduce the frequency of video decoding and accelerate the loading performance.

% \textbf{Compression Loss vs. Size Reduction}: algname is a versatile format that supports multiple compression and encoding methods. The impact of lossy compression varies depending on the specific task and compression scheme. Future work will explore minimizing the effects of compression on training results.

% \newpage 
% \bibliographystyle{IEEEtran}
% \bibliography{IEEEabrv,references}

\renewcommand*{\bibfont}{\footnotesize}
\printbibliography

@STRING{icra = {Proc. {IEEE} Int. Conf. Robotics and Automation (ICRA)}}

@STRING{iros = {Proc. IEEE/RSJ Int. Conf. on Intelligent Robots and Systems (IROS)}}

@STRING{case = {Proc. {IEEE} Conf. on Automation Science and Engineering (CASE)}}

@STRING{rss = {Proc. Robotics: Science and Systems (RSS)}}

@STRING{corl = {Conf. on Robot Learning (CoRL)}}

@misc{mkv,
  title = {{Matroska Video Container}},
  howpublished = {\url{https://www.matroska.org/index.html}},
  note = {Accessed: 2024-09-14}
}

@misc{arrow,
  title = {{Apache Arrow}},
  howpublished = {\url{https://arrow.apache.org/}},
  note = {2024-09-14}
}

@misc{decord,
  title = {{Decord: An efficient video loader for deep learning with smart shuffling that's super easy to digest}},
  howpublished = {\url{https://github.com/dmlc/decord}},
  note = {Accessed: 2024-09-14}
}

@inproceedings{sagiroglu2013big,
  title={Big data: A review},
  author={Sagiroglu, Seref and Sinanc, Duygu},
  booktitle={2013 international conference on collaboration technologies and systems (CTS)},
  pages={42--47},
  year={2013},
  organization={IEEE}
}

@misc{nvdec,
  title = {{NVIDIA Video Codec SDK}},
  howpublished = {\url{https://developer.nvidia.com/video-codec-sdk}},
  note = {Accessed: 2024-09-14}
}

@misc{mmap,
  title = {{Linux mmap(2) Manual}},
  howpublished = {\url{https://man7.org/linux/man-pages/man2/mmap.2.html}},
  note = {Accessed: 2024-09-14}
}

@misc{pyav,
  title = {{Pyav: Pythonic bindings for FFmpeg's libraries}},
  howpublished = {\url{https://github.com/PyAV-Org/PyAV}},
  note = {Accessed: 2024-09-14}
}

@misc{safetensors,
  title = {{HuggingFace SafeTensors}},
  howpublished = {\url{https://github.com/huggingface/safetensors}},
  note = {Accessed: 2024-09-14}
}

@misc{lerobot_benchmark,
  title = {{LeRobot Video Benchmark}},
  howpublished = {\url{https://github.com/huggingface/lerobot/tree/main/benchmarks/video}},
  note = {Accessed: 2024-09-13}
}

@inproceedings{ichnowski2020fog,
  title={Fog Robotics Algorithms for Distributed Motion Planning Using Lambda Serverless Computing},
  author={Ichnowski, Jeffrey and Lee, William and Murta, Victor and Paradis, Samuel and Alterovitz, Ron and Gonzalez, Joseph E and Stoica, Ion and Goldberg, Ken},
  booktitle=icra,
  pages={4232--4238},
  year={2020}
}

@inproceedings{tanwani2019fog,
  title={A fog robotics approach to deep robot learning: Application to object recognition and grasp planning in surface decluttering},
  author={Tanwani, Ajay Kumar and Mor, Nitesh and Kubiatowicz, John and Gonzalez, Joseph E and Goldberg, Ken},
  booktitle=icra,
  pages={4559--4566},
  year={2019},
  organization={IEEE}
}

@inproceedings{quigley2009ros,
  title={{ROS}: an open-source Robot Operating System},
  author={Quigley, Morgan and Gerkey, Brian and Conley, Ken and Faust, Josh and Foote, Tully and Leibs, Jeremy and Wheeler, Rob and Ng, Andrew},
  booktitle={ICRA workshop on open source software},
  volume={3},
  number={3.2},
  year={2009}
}

@inproceedings{chen2021fogros,
  title={{FogROS}: An Adaptive Framework for Automating Fog Robotics Deployment},
  author={Chen, Kaiyuan Eric and Liang, Yafei and Jha, Nikhil and Ichnowski, Jeffrey and Danielczuk, Michael and Gonzalez, Joseph and Kubiatowicz, John and Goldberg, Ken},
  booktitle={2021 IEEE 17th International Conference on Automation Science and Engineering (CASE)},
  pages={2035--2042},
  year={2021},
  organization={IEEE}
}

@misc{foxglove,
  title = {{Foxglove}},
  author = {{Foxglove Technologies Inc}},
  howpublished={\url{https://foxglove.dev}}
}

@inproceedings{tian2019fog,
  title={A fog robotic system for dynamic visual servoing},
  author={Tian, Nan and Tanwani, Ajay Kummar and Chen, Jinfa and Ma, Mas and Zhang, Robert and Huang, Bill and Goldberg, Ken and Sojoudi, Somayeh},
  booktitle={2019 International Conference on Robotics and Automation (ICRA)},
  pages={1982--1988},
  year={2019},
  organization={IEEE}
}

@inproceedings{gudi2017fog,
  title={Fog robotics: An introduction},
  author={Gudi, Siva Leela Krishna Chand and Ojha, Suman and Johnston, Benjamin and Clark, Jesse and Williams, Mary-Anne},
  booktitle={IEEE/RSJ International Conference on Intelligent Robots and Systems},
  year={2017}
}

@inproceedings{gudi2018fog,
  title={Fog robotics for efficient, fluent and robust human-robot interaction},
  author={Gudi, Siva Leela Krishna Chand and Ojha, Suman and Johnston, Benjamin and Clark, Jesse and Williams, Mary-Anne},
  booktitle={2018 IEEE 17th International Symposium on Network Computing and Applications (NCA)},
  pages={1--5},
  year={2018},
  organization={IEEE}
}

@article{chen2023sgc,
  title={{FogROS2-SGC}: A {ROS2} Cloud Robotics Platform for Secure Global Connectivity},
  author={Kaiyuan Chen and Ryan Hoque and K Dharmarajan and Edith Llontop and S. O. Adebola and Jeffrey Ichnowski and John D. Kubiatowicz and Ken Goldberg},
  journal={2023 IEEE/RSJ International Conference on Intelligent Robots and Systems (IROS)},
  year={2023},
  pages={1-8},
  url={https://api.semanticscholar.org/CorpusID:259287275}
}

@article{chen2024fogrosls,
  title={{FogROS2-LS}: {A} Location-Independent Fog Robotics Framework for Latency Sensitive {ROS2} Applications},
  author={Chen, Kaiyuan and Wang, Michael and Gualtieri, Marcus and Tian, Nan and Juette, Christian and Ren, Liu and Kubiatowicz, John and  Goldberg, Ken},
  journal=icra,
  year={2024},
  publisher={IEEE}
}

@article{chen2024fogrosconfig,
  title={{FogROS2-Config}: {A} Toolkit for Choosing Server Configuration For Cloud Robotics},
  author={Chen, Kaiyuan and Hari, Kush and Khare, Rohil and Le, Charlotte and Chung, Trinity and Drake, Jaimyn and Adebloa, Simeon and Ichnowski, Jeffrey and Kubiatowicz, John and Goldberg, Ken},
  journal=icra,
  year={2024},
  publisher={IEEE}
}

@article{brohan2023rt1,
  title={{RT}-1: Robotics transformer for real-world control at scale},
  author={Anthony Brohan and Noah Brown and Justice Carbajal and Yevgen Chebotar and Chelsea Finn and Keerthana Gopalakrishnan and Karol Hausman and Alexander Herzog and Jasmine Hsu and Brianna Zitkovich and Others},
  journal={Robotics: Science and Systems (RSS)},
  year={2023}
}

@inproceedings{brohan2023rt2,
  title={Rt-2: Vision-language-action models transfer web knowledge to robotic control},
  author={Anthony Brohan and Noah Brown and Justice Carbajal and Yevgen Chebotar and Chelsea Finn and Keerthana Gopalakrishnan and Karol Hausman and Alexander Herzog and Jasmine Hsu and Brianna Zitkovich and Others},
  booktitle={Conference on Robot Learning},
  pages={2165--2183},
  year={2023},
  organization={PMLR}
}

@inproceedings{kalashnikov2018qt,
  title={Scalable deep reinforcement learning for vision-based robotic manipulation},
  author={Dmitry Kalashnikov and Alex Irpan and Peter Pastor and Julian Ibarz and Alexander Herzog and Eric Jang and Deirdre Quillen and Ethan Holly and Mrinal Kalakrishnan and Vincent Vanhoucke and Sergey Levine},
  booktitle={Conference on robot learning},
  pages={651--673},
  year={2018},
  organization={PMLR}
}

@misc{octo_2023,
    title={Octo: An Open-Source Generalist Robot Policy},
    author = {{Octo Model Team} and Dibya Ghosh and Homer Walke and Karl Pertsch and Kevin Black and Oier Mees and Sudeep Dasari and Joey Hejna and Charles Xu and Jianlan Luo and Tobias Kreiman and {You Liang} Tan and Dorsa Sadigh and Chelsea Finn and Sergey Levine},
    howpublished  = {\url{https://octo-models.github.io}},
    year = {2023},
}

@misc{open_x_embodiment_rt_x_2023,
title={Open {X-E}mbodiment: Robotic Learning Datasets and {RT-X} Models},
author = {Open X-Embodiment Collaboration and Others},
year = {2024},
}

@article{alayrac2022flamingo,
  title={Flamingo: a visual language model for few-shot learning},
  author={Jean-Baptiste Alayrac and Jeff Donahue and Pauline Luc and Antoine Miech and Iain Barr and Yana Hasson and Karel Lenc and Arthur Mensch and Katie Millican and Malcolm Reynolds and Roman Ring and Eliza Rutherford and Serkan Cabi and Tengda Han and Zhitao Gong and Sina Samangooei and Marianne Monteiro and Jacob Menick and Sebastian Borgeaud and Andrew Brock and Aida Nematzadeh and Sahand Sharifzadeh and Mikolaj Binkowski and Ricardo Barreira and Oriol Vinyals and Andrew Zisserman and Karen Simonyan},
  journal={Advances in neural information processing systems},
  volume={35},
  pages={23716--23736},
  year={2022}
}

@inproceedings{driess2023palme,
  title={PaLM-E: An Embodied Multimodal Language Model},
  author={Danny Driess and Fei Xia and Mehdi S. M. Sajjadi and Corey Lynch and Aakanksha Chowdhery and Brian Ichter and Ayzaan Wahid and Jonathan Tompson and Quan Vuong and Tianhe Yu and Wenlong Huang and Yevgen Chebotar and Pierre Sermanet and Daniel Duckworth and Sergey Levine and Vincent Vanhoucke and Karol Hausman and Marc Toussaint and Klaus Greff and Andy Zeng and Igor Mordatch and Pete Florence},
  booktitle={International Conference on Machine Learning},
  pages={8469--8488},
  year={2023},
  organization={PMLR}
}

@inproceedings{walke2023bridgedata,
  title={Bridgedata v2: A dataset for robot learning at scale},
  author={Homer Walke and Kevin Black and Abraham Lee and Moo Jin Kim and Max Du and Chongyi Zheng and Tony Zhao and Philippe Hansen-Estruch and Quan Vuong and Andre He and Vivek Myers and Kuan Fang and Chelsea Finn and Sergey Levine},
  booktitle={Conference on Robot Learning},
  pages={1723--1736},
  year={2023},
  organization={PMLR}
}

@misc{BerkeleyUR5Website,
  title = {Berkeley {UR5} Demonstration Dataset},
  author = {Lawrence Yunliang Chen and Simeon Adebola and Ken Goldberg},
  howpublished = {https://sites.google.com/view/berkeley-ur5/home},
}

@inproceedings{khazatsky2024droid,
    title   = {DROID: A Large-Scale In-The-Wild Robot Manipulation Dataset},
    author  = {Alexander Khazatsky and Karl Pertsch and Suraj Nair and Ashwin Balakrishna and Sudeep Dasari and Siddharth Karamcheti and Soroush Nasiriany and Mohan Kumar Srirama and Lawrence Yunliang Chen and Others},
    booktitle = {Proceedings of Robotics: Science and Systems},
    address  = {Delft, Netherlands},
    year = {2024},
}

@article{achiam2023gpt4,
  title={Gpt-4 technical report},
  author={Achiam, Josh and Adler, Steven and Agarwal, Sandhini and Ahmad, Lama and Akkaya, Ilge and Aleman, Florencia Leoni and Almeida, Diogo and Altenschmidt, Janko and Altman, Sam and Anadkat, Shyamal and others},
  journal={arXiv preprint arXiv:2303.08774},
  year={2023}
}

@article{touvron2023llama,
  title={Llama: Open and efficient foundation language models},
  author={Hugo Touvron and Thibaut Lavril and Gautier Izacard and Xavier Martinet and Marie-Anne Lachaux and Timothée Lacroix and Baptiste Rozière and Naman Goyal and Eric Hambro and Faisal Azhar and Aurelien Rodriguez and Armand Joulin and Edouard Grave and Guillaume Lample},
  journal={arXiv preprint arXiv:2302.13971},
  year={2023}
}

@article{liu2024visual,
  title={Visual instruction tuning},
  author={Liu, Haotian and Li, Chunyuan and Wu, Qingyang and Lee, Yong Jae},
  journal={Advances in neural information processing systems},
  volume={36},
  year={2024}
}

@article{fu2024icrt,
    title={In-Context Imitation Learning via Next-Token Prediction}, 
    author={Letian Fu and Huang Huang and Gaurav Datta and Lawrence Yunliang Chen and William Chung-Ho Panitch and Fangchen Liu and Hui Li and Ken Goldberg},
    journal={International Conference on Robotics and Automation},
    year={2025}
}

@article{kim24openvla,
    title={OpenVLA: An Open-Source Vision-Language-Action Model},
    author={{Moo Jin} Kim and Karl Pertsch and Siddharth Karamcheti and Ted Xiao and Ashwin Balakrishna and Suraj Nair and Rafael Rafailov and Ethan Foster and Grace Lam and Pannag Sanketi and Quan Vuong and Thomas Kollar and Benjamin Burchfiel and Russ Tedrake and Dorsa Sadigh and Sergey Levine and Percy Liang and Chelsea Finn},
    journal = {Conference on Robot Learning (CoRL)},
    year={2024},
}

@inproceedings{2023GPT4VisionSC,
  title={GPT-4V(ision) System Card},
  author={},
  year={2023},
  url={https://api.semanticscholar.org/CorpusID:263218031}
}

@article{2023gemini,
  title={Gemini: A Family of Highly Capable Multimodal Models},
  author={Google},
  journal={arXiv preprint arXiv:2312.11805},
  year={2023}
}

@misc{cadene2024lerobot,
  author = {Cadene, Remi and Alibert, Simon and Soare, Alexander and Gallouedec, Quentin and Wolf, Thomas},
  title = {LeRobot: Making AI for Robotics more accessible with end-to-end learning},
  year = {2024},
  publisher = {GitHub},
  journal = {GitHub repository},
  howpublished = {\url{https://github.com/huggingface/lerobot}}
}

@software{hdf5,
  author = {{The HDF Group}},
  title = {{Hierarchical Data Format, version 5}},
  year = {1997-2024},
  url = {https://www.hdfgroup.org/HDF5/},
}

@article{hussenot2021rlds,
  title={RLDS: an Ecosystem to Generate, Share and Use Datasets in Reinforcement Learning},
  author={Sabela Ramos and Sertan Girgin and Léonard Hussenot and Damien Vincent and Hanna Yakubovich and Daniel Toyama and Anita Gergely and Piotr Stanczyk and Raphael Marinier and Jeremiah Harmsen and Olivier Pietquin and Nikola Momchev},
  journal={arXiv preprint arXiv:2111.02767},
  year={2021}
}

@techreport{h264,
  title = {Advanced video coding for generic audiovisual services},
  author = {{ITU-T}},
  year = {2003},
  institution = {International Telecommunication Union},
  type = {Recommendation},
  number = {H.264},
  address = {Geneva, Switzerland}
}

@techreport{h265,
  title = {High efficiency video coding},
  author = {{ITU-T}},
  year = {2023},
  institution = {International Telecommunication Union},
  type = {Recommendation},
  number = {H.265},
  address = {Geneva, Switzerland},
  note = {Version 9}
}

@misc{av1,
  title = {AV1 Bitstream \& Decoding Process Specification},
  author = {{Alliance for Open Media}},
  year = {2019},
  howpublished = {\url{https://aomediacodec.github.io/av1-spec/}},
  note = {Accessed: [Insert Date]}
}

@misc{loc_ffv1,
  title = {FF Video Codec 1, Version 0, 1 and 3},
  author = {{Library of Congress}},
  year = {2024},
  howpublished = {\url{https://www.loc.gov/preservation/digital/formats/fdd/fdd000341.shtml}},
  note = {Accessed: [Insert Date]}
}

@techreport{rfc9043,
  author = {M. Niedermayer and D. Rice and J. Martinez},
  title = {FFV1 Video Coding Format Versions 0, 1, and 3},
  howpublished = {Internet Requests for Comments},
  type = {RFC},
  number = {9043},
  year = {2021},
  month = {August},
  issn = {2070-1721},
  publisher = {RFC Editor},
  institution = {RFC Editor},
  url = {https://www.rfc-editor.org/info/rfc9043}
}

@techreport{rfc8794,
  author = {S. Lhomme and D. Rice and M. Bunkus},
  title = {Extensible Binary Meta Language},
  howpublished = {Internet Requests for Comments},
  type = {RFC},
  number = {8794},
  year = {2020},
  month = {July},
  issn = {2070-1721},
  publisher = {RFC Editor},
  institution = {RFC Editor},
  url = {https://www.rfc-editor.org/info/rfc8794}
}

@article{kam2015rviz,
  title={RViz: a toolkit for real domain data visualization},
  author={Kam, Hyeong Ryeol and Lee, Sung-Ho and Park, Taejung and Kim, Chang-Hun},
  journal={Telecommunication Systems},
  volume={60},
  number={2},
  pages={337--345},
  year={2015},
  publisher={Springer}
}

@article{luo2024multi,
  title={Multi-stage cable routing through hierarchical imitation learning},
  author={Luo, Jianlan and Xu, Charles and Geng, Xinyang and Feng, Gilbert and Fang, Kuan and Tan, Liam and Schaal, Stefan and Levine, Sergey},
  journal={IEEE Transactions on Robotics},
  year={2024},
  publisher={IEEE}
}

@article{pari2021surprising,
    title={The Surprising Effectiveness of Representation Learning for Visual Imitation}, 
    author={Jyothish Pari and Nur Muhammad Shafiullah and Sridhar Pandian Arunachalam and Lerrel Pinto},
    year={2018},
    journal={Robotics: Science and Systems}
}

@misc{TFDS,
  title = {{TensorFlow Datasets}, A collection of ready-to-use datasets},
  howpublished = {\url{https://www.tensorflow.org/datasets}},
}
\end{document}